\documentclass{article}
\usepackage[preprint]{spconf}
\usepackage{amsmath,graphicx,booktabs,url}
\usepackage[utf8]{inputenc}
\usepackage[T1]{fontenc}

\title{Self-Supervised Representation Learning:\\
From Spectral Foundation Models to Auroral Emission Spectra}

\name{Matthieu Le Lain$^{1}$ \qquad Ga\"el Cessateur$^{2}$ \qquad
      S\'ebastien Lef\`evre$^{1,3}$}
\address{$^{1}$IRISA, UMR CNRS 6074, Universit\'e Bretagne Sud, Vannes, France\\
         $^{2}$Royal Belgian Institute for Space Aeronomy, Brussels, Belgium\\
         $^{3}$UiT - The Arctic University of Norway, Troms\o{}, Norway}

\ninept
\renewcommand{\footnotesize}{\small}
\begin{document}
\toappear{\parbox[b]{\textwidth}{\centering\footnotesize This work has been submitted to the IEEE for possible publication. Copyright may be transferred without notice, after which this version may no longer be accessible.}}
\maketitle

\begin{abstract}
Auroral spectrographs such as the Auroral Spectrograph In Skibotn (ASIS)
record hundreds of thousands of emission spectra, but only a few hundred
can be labelled by an expert. To exploit the rest, we pretrain a 1D Vision
Transformer with a masked autoencoder on 223\,000 unlabelled spectra.
Without labels, its representation recovers the emission-line intensity
ratios that physicists use to diagnose the precipitating particles (R$^2$
0.91 vs.\ 0.77 for an untrained control) and, under one linear probe,
classifies as well as 13 features designed by experts. Fine-tuned, the
model outperforms the previous supervised auroral classifier on its own
benchmark (macro-AP 88.5 vs.\ 77.8), reaches 0.870 mAP, and exceeds the same architecture
trained from scratch by $+0.159$ with 10\,\% of the labels; attribution shows that it uses both N$_2^+$
bands. Could an existing pretrained model replace it? Two astronomical spectral foundation models and a time-series
model transfer according to their spectral window: SpectraFM, trained in
the infrared, falls below the untrained control, whereas SpecFormer,
trained in the optical, approaches in-domain pretraining without reaching
it.
\end{abstract}


\section{Introduction}
\label{sec:intro}

An auroral emission spectrum records which atoms and molecules of the
upper atmosphere are excited by precipitating particles, and at which
energy. Line intensities and their ratios are established diagnostics of
this precipitation: the 630/557.7\,nm ratio of the two forbidden oxygen
lines traces the energy of precipitating electrons~\cite{rees1974auroral},
and the hydrogen Balmer lines signal proton
precipitation~\cite{eather1967auroral}. The ASIS spectrograph at Skibotn,
Norway~\cite{Cessateur2026}, has recorded nearly 330\,000 calibrated spectra
over 220 nights since 2023. Exploiting such a corpus requires knowing, for
each spectrum, which signatures it contains, and today this is decided by
an expert who inspects the spectra one by one: after three observing
seasons, only 811 spectra have been labelled.

Two routes can exploit the unlabelled corpus. The first is
self-supervised learning (SSL): pretraining an encoder on the spectra
themselves, so that a classifier trained afterwards needs far fewer
labels. The second is to reuse a spectral foundation model pretrained
elsewhere, on stellar or galaxy spectra. Whether such models, built for
the absorption spectra of astronomical sources, transfer to the emission
spectra of the upper atmosphere has never been measured. The answer decides whether a new spectroscopic domain must pretrain its own
encoder or can reuse an existing one: what transfers, and why?

Our contributions are twofold. (1)~A self-supervised encoder for
auroral spectra whose frozen representation, learned without labels,
recovers these line ratios and, under the same linear probe, classifies as
well as 13 hand-crafted features that encode the diagnostics an expert reads
on a spectrum (line intensities, their ratios and the continuum level); fine-tuned, it beats the same architecture
trained from scratch at every label budget and outperforms the previous
supervised auroral classifier~\cite{vit1d2026} on its own benchmark, and
attribution shows that it relies on physically coherent emission systems.
(2)~Under one evaluation protocol applied identically to every
representation, we compare this in-domain pretraining with two astronomical
foundation models, SpectraFM~\cite{koblischke2024spectrafm} (infrared
stellar spectra) and SpecFormer~\cite{parker2024astroclip} (optical galaxy
spectra), and with a generic time-series model,
Mantis~\cite{feofanov2025mantis}: none replaces in-domain pretraining, and
their transfer follows the overlap between their pretraining wavelength
window and ours rather than the physical nature of the sources.

Machine learning for aurorae has focused on all-sky images, classified by
morphology~\cite{syrjasuo2004,clausen2018,kvammen2020auroral,johnson2024themis};
spectra, which resolve the emitting species, have received little
attention. Our earlier supervised 1D Vision Transformer
(ViT-1D)~\cite{vit1d2026}, trained on 719 labelled spectra, had to use 8
coarse tokens of 35\,nm to avoid overfitting and did not outperform a
single-patch MLP (macro-AP 77.8 vs.\ 81.1). Masked autoencoders (MAE)~\cite{he2022mae}
have been applied to 1D signals, from Raman spectra~\cite{raman2025mae}
to electrocardiograms~\cite{ecg2023maskedtransformer}; autoencoders have been
trained on astronomical spectra~\cite{melchior2023spender,sedaghat2021machines}, and spectral
foundation models now exist: SpectraFM~\cite{koblischke2024spectrafm},
trained on APOGEE infrared stellar spectra, a stellar Transformer trained on
Gaia XP spectra~\cite{leung2023towards}, SpecFormer, the spectrum encoder of
AstroCLIP~\cite{parker2024astroclip}, trained on DESI optical galaxy
spectra, and OmniSpectra~\cite{omnispectra2026}. Informed
masking, which concentrates the mask on informative regions, is
established in
vision~\cite{bandara2023adamae,kakogeorgiou2022attmask,li2022semmae,wang2023hpm},
but adapts the mask to each image, whereas in a spectrum the informative
positions are fixed along the wavelength axis, which allows a static prior.
Mantis~\cite{feofanov2025mantis}, a generic time-series foundation
model, provides a comparison point with no spectral prior. None of these
works evaluates transfer to atmospheric emission spectra, where lines are
in emission, there is no redshift, the continuum is a nuisance, and the
classes are combinations of lines.
\section{Data}
\label{sec:data}

The corpus contains 329\,704 calibrated spectra recorded by ASIS over 220
nights between 2023 and 2026. Each spectrum covers 405--687\,nm
in 1024 points (0.28\,nm each). Consecutive spectra of the same night are near-duplicates, so the corpus is
split by night: 154 nights (about 230k spectra) are used for pretraining, 33 validation
and 33 test nights (about 50k and 49k spectra) are held out. Spectra with saturated, null or non-finite pixels are left out of
training (97.2\,\% pass). After this
filter and the removal of the labelled spectra, pretraining uses 223\,065
spectra.

An auroral spectrum is the sum of narrow emission lines, which carry the
physics, and a smooth continuum from scattered light (Moon, twilight,
clouds), whose brightness varies far more than the lines and would
dominate any representation. We therefore give the model two input
channels. i) The line channel is the spectrum minus a running-median baseline
(window of 101 points, Gaussian-smoothed with $\sigma=8$ points), clipped
at zero, compressed by an asinh transform and standardised per spectrum, with 8
points masked at each edge; ii) the continuum channel is the baseline itself,
asinh-compressed and downsampled by 8. A median baseline is preferred to a
low percentile because it stays unbiased on the sloped continuum of
twilight (low-frequency residual left in the line channel, 90th percentile
across spectra: 0.22, against 0.52 with a 10th-percentile baseline). The separate continuum channel
preserves the scattered-light information that two classes below rely on.

An expert labelled 811 spectra from six nights with seven classes
(Fig.~\ref{fig:interp}a). Each
class marks the presence of one signature, and a spectrum can carry
several labels because several mechanisms are often active at once, for
instance proton and electron precipitation, which makes the task
multilabel: \texttt{ha} (H$\alpha$ 656.3\,nm, proton precipitation; 557
positives), \texttt{elow} (high 630/557.7 ratio, low-energy electrons;
214), \texttt{n2\_471} (N$_2^+$ first-negative band at 470.9\,nm,
energetic electrons; 267), \texttt{molecular}
(N$_2$ first-positive bands, 654--687\,nm; 326), \texttt{o\_520} (the [N\,I] line at 520.0\,nm; 307),
and two classes that flag spectra dominated by scattered light rather than
aurora: \texttt{cloudy} (clouds or Moon; 531) and \texttt{dawn\_dusk} (twilight,
with the sodium D line at 589.2\,nm; 27). The 719 spectra labelled with
five classes in~\cite{vit1d2026} are a subset of these 811 and serve as
the benchmark against the previous supervised model.

\section{Method}
\label{sec:method}

\subsection{Encoder and pretraining}
The encoder is a 1D Vision Transformer (ViT-1D) with two input channels. A
convolution with a 16-point patch turns the 1024 points into 64 tokens;
each token spans 4.4\,nm, about the width of one emission line, so that a
token corresponds to one physical feature. The tokens are processed by 8
Transformer blocks of width 256 with 8 attention heads (6.3\,M
parameters). Such fine tokens overfit with 719 labels~\cite{vit1d2026};
pretraining on 223\,065 unlabelled spectra removes this issue.

Pretraining is performed with a standard MAE~\cite{he2022mae} adapted to two
channels: 35\,\% of the tokens are masked uniformly at random, the encoder
sees only the visible tokens of both channels, and a light decoder (width 128, 3 blocks,
discarded after pretraining) reconstructs the masked tokens; the loss is
the mean squared error on the masked tokens, averaged over the two channels
with equal weights (150 epochs, batch 512). The 811 labelled spectra are excluded from
pretraining, whose objective uses no label; labels serve only for model
selection, as a validation criterion: every ten epochs, a linear classifier
is fitted on the frozen embeddings of the 811 labelled spectra, and the
checkpoint with the best score is kept, for every pretrained variant alike. The probe score peaks at epoch 20 and declines by 0.02
by epoch 150 while the reconstruction loss keeps decreasing.

As an ablation, we also test an informed prior as presented in Section~\ref{ssec:mask}:
uniform random masking spends most of the masking budget on the featureless continuum between lines. Since the informative positions of a spectrum are
fixed along the wavelength axis, we compute once, before training, the
standard deviation of the line channel at each of the 64 token positions
over a 20\,000-spectrum sample of the corpus. This uses no catalogue and no
label, yet its four largest values fall on the tokens of the brightest auroral
lines, [O I] 557.7, the red oxygen doublet [O I] 630.0 and 636.4, and NaD
589.2\,nm (all in the catalogue of Sec.~\ref{ssec:protocol}), and the
N$_2^+$ 427.8\,nm token ranks seventh: the prior
rediscovers the brightest auroral lines. Tokens are then masked with probability proportional to
this information (possibly square-rooted), at a constant
global masking rate, by Gumbel top-$k$ sampling without
replacement~\cite{kool2019gumbel}.

\subsection{Evaluation protocol}
\label{ssec:protocol}
The labelled spectra are evaluated by stratified multilabel 5-fold
cross-validation repeated 3 times (15 folds), with the same folds, seeds
and label subsets for every representation and model; standardisation,
decision thresholds and regularisation are fitted on the training folds
only. Because spectra of the same night fall in both training and test
folds, the scores measure generalisation to unseen spectra, not to unseen
nights. A frozen representation is scored by a linear probe: one logistic
regression per class, whose L2 regularisation strength is selected by
3-fold inner cross-validation on the training folds, since a single fixed
strength would favour 13 features over 512- to 768-value embeddings, and
whose decision threshold maximises F1 on the training folds (a default
threshold penalises rare classes). The main metric
is the average precision averaged over the seven classes (macro mAP),
which weights the rare class \texttt{dawn\_dusk} like the others; macro F1 and AUROC complement it, and few-shot curves repeat
the evaluation with 10, 25, 50 and 100\,\% of the training labels.
Differences are reported as folds won out of 15 by the first model, with
a paired Wilcoxon $p$-value (indicative only: the three repetitions share
spectra).

Two references frame every score. The upper reference is the expert
baseline: 13 hand-crafted features encoding the specialists' diagnostics,
namely the intensities of 7 lines of the auroral
catalogue~\cite{vallancejones1974aurora} (N$_2^+$ 427.8, [N I] 520.0,
[O I] 557.7, NaD 589.2, [O I] 630.0, [O I] 636.4 and H$\alpha$ 656.3\,nm),
each measured as the radiance above the spectrum's 5th-percentile level
within $\pm0.6$\,nm of the line; 4 logarithmic ratios of these intensities
to the 557.7 line, the classical precipitation
diagnostics, plus the 636.4/630 ratio of the red oxygen doublet, fixed by atomic
physics (both lines share the same upper level) and used as a control; and the continuum level. The lower reference is a
control encoder with the same architecture and random, untrained weights:
what a pretrained representation scores above it is what pretraining has
added.

\subsection{Fine-tuning}
For classification, the encoder output (the class token concatenated with
the mean of the 64 tokens, 512 values) feeds a linear layer with seven outputs. Training minimises a
class-weighted binary cross-entropy (\texttt{dawn\_dusk} is present in
3\,\% of the spectra), adds Gaussian noise ($\sigma{=}0.05$) as
augmentation, and uses AdamW with learning rates of $10^{-3}$ for the head
and $10^{-4}$ for the encoder under a cosine schedule, for 15 epochs (100
when only the head is trained). The
from-scratch control trains the same
architecture with the same recipe from random weights: only the
initialisation differs, so the gap between the two is attributable to
pretraining alone.

\section{Results}
\label{sec:results}

\begin{figure*}[t]
\centering
\includegraphics[width=\textwidth]{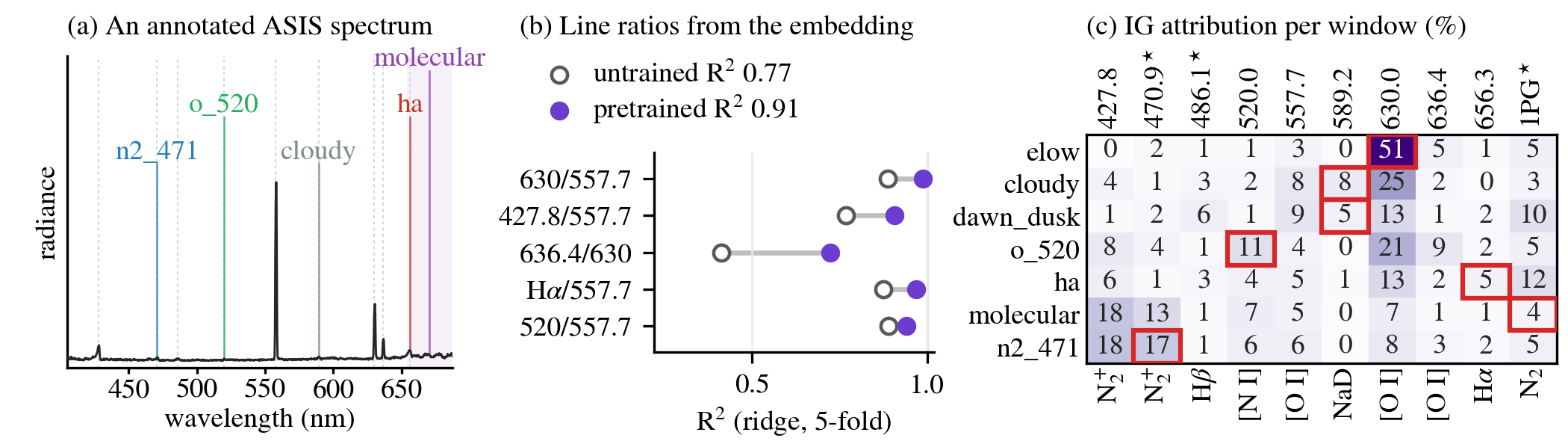}
\caption{(a) An ASIS spectrum, the expert's classes above their criterion. (b) Ridge-regression R$^2$ of five line ratios read from the frozen
embedding, pretrained encoder vs.\ untrained control (5-fold; ratios clipped
at the 1st/99th percentiles; legend: mean of the five). (c) Share of positive IG attribution per window ($\pm2$\,nm around nine lines; N$_2$
first-positive (1PG) band: 654--687\,nm minus the H$\alpha$ window), per class of the fine-tuned classifier on its positives; rest outside the
windows; red frame = annotation criterion; $\star$ = absent from the expert
features.}
\label{fig:interp}
\end{figure*}

\subsection{Transfer of frozen representations}

\begin{table}[t]
\centering
\caption{Frozen representations under one linear probe (logistic regression,
regularisation selected by inner cross-validation on the training folds),
7 classes, $5{\times}3$ cross-validation, $\pm$ standard deviation across
folds. Window: wavelengths of the pretraining data. Last column: mAP with
10\,\% of the annotations (subsets of Table~\ref{tab:seven}). +: concatenation of the two. Bold: best single representation.}
\label{tab:transfer}
\vspace{2pt}
\small
\setlength{\tabcolsep}{2pt}\begin{tabular}{lccc}
\toprule
Representation & Window (nm) & mAP & @10\,\% \\
\midrule
\emph{Untrained control} & --- & 0.816{$\pm$0.034} & 0.683{$\pm$0.036} \\
Expert features & --- & 0.861{$\pm$0.032} & 0.823{$\pm$0.041} \\
Ours (auroral MAE) & 405--687 & \textbf{0.863}{$\pm$0.037} & \textbf{0.829}{$\pm$0.036} \\
Expert feat.\ + Ours & 405--687 & 0.864{$\pm$0.038} & 0.831{$\pm$0.037} \\
SpecFormer (galaxies) & 360--980 & 0.844{$\pm$0.028} & 0.805{$\pm$0.037} \\
Mantis (time series) & --- & 0.824{$\pm$0.025} & 0.758{$\pm$0.047} \\
SpectraFM (stars) & 1515--1694 & 0.789{$\pm$0.033} & 0.701{$\pm$0.049} \\
\bottomrule
\end{tabular}
\end{table}

Table~\ref{tab:transfer} gives the linear-probe macro mAP with all labels,
then with 10\,\%. The untrained encoder scores 0.816, which sets the reference floor. Our
in-domain pretraining reaches 0.863, $+0.047$ over the floor (14 of 15
folds, $p<0.001$), and ties with the expert features (0.861; $p=0.76$): a
representation learned from raw spectra, with no line catalogue, carries the same information as the 13 hand-designed diagnostics. Concatenating the two yields no gain (0.864, $p=0.30$; Table~\ref{tab:transfer}): the encoder is redundant with the features rather than complementary to them.

Among the external models, transfer follows the spectral window. SpectraFM
and SpecFormer are both astronomical foundation models; their decisive
difference is their wavelength range.
SpectraFM was trained in the infrared (1515--1694\,nm), with no overlap
with our 405--687\,nm: it falls below the untrained control (0.789, $p=0.005$), so a spectral prior learned in the wrong window is
worse than no spectral prior at all. Mantis, pretrained on generic time series only, does not rise above the floor (0.824, $p=0.56$). SpecFormer was
trained on galaxy spectra in an optical window that contains ours
(360--980\,nm): it is the only external model above the floor (0.844, $p=0.01$), yet it stays below in-domain pretraining ($p=0.03$). Transfer therefore depends on sharing the measurement window rather than
on the physical proximity of the sources. With 10\,\%
of the labels the order is unchanged, except that SpectraFM no longer
differs from the floor (0.701 vs.\ 0.683, $p=0.25$). External models are used frozen (spectral ones: best of three input scalings).

\subsection{Five-class benchmark against the supervised ViT-1D}
The previous supervised ViT-1D~\cite{vit1d2026} was evaluated on 719
spectra with five classes; we reuse its labels, cross-validation scheme
and seed. Our
fine-tuned model reaches a macro-AP of 88.5 against 81.1 for the MLP and
77.8 for the ViT-1D of that work, and a macro-F1 of 82.7 against 74.4 and
73.0; all five classes improve, most of all the hardest, \texttt{molecular},
from 51.6 to 82.5. Part of this gain comes from our pipeline rather than
from pretraining: the same ViT-1D trained from scratch in our pipeline
already reaches 83.0, so pretraining accounts for the remaining $+5.5$,
concentrated on \texttt{molecular} (62.6 $\to$ 82.5). The expert features,
probed on the same folds, reach 86.5: the fine-tuned model exceeds them by
$+2.0$ (14 of 15 folds, $p<0.001$), the one setting of this study where the
gap to expert knowledge is significant.

\subsection{Seven-class multilabel task}
\label{ssec:seven}

\begin{table}[t]
\centering
\caption{Seven-class fine-tuning (811 spectra, $5{\times}3$ cross-validation). Top:
training strategies (mAP, F1, AUROC), all with the same linear head (frozen,
head only: encoder not updated); the expert-feature row is the probe of
Table~\ref{tab:transfer}. Bottom: label efficiency (mAP)
with a fraction of each fold's training labels, same subsets for the three
columns; last column, gain of pretraining over the same model from scratch.
Bold: best per column (top) and row (bottom).}
\label{tab:seven}
\vspace{2pt}
\small
\begin{tabular}{lccc}
\toprule
Configuration & mAP & F1 & AUROC \\
\midrule
Expert features (probe) & 0.861 {$\pm$0.032} & 0.789 & \textbf{0.910} \\
SSL frozen, head only & \textbf{0.870} {$\pm$0.034} & \textbf{0.812} & 0.908 \\
SSL fine-tuned, 15 ep. & \textbf{0.870} {$\pm$0.032} & 0.804 & 0.909 \\
SSL fine-tuned, 100 ep. & 0.839 {$\pm$0.030} & 0.731 & 0.890 \\
\midrule
\hspace{1mm}From scratch, 15 ep. & 0.840 {$\pm$0.033} & 0.776 & 0.877 \\
\hspace{1mm}From scratch, 100 ep. & 0.800 {$\pm$0.037} & 0.730 & 0.856 \\
\hspace{1mm}Untrained, head only & 0.732 {$\pm$0.029} & 0.703 & 0.796 \\
\bottomrule
\end{tabular}

\vspace{5pt}
\begin{tabular}{lcccc}
\toprule
Labels & Expert feat. & From scratch & \textbf{SSL} & $\Delta$ \\
\midrule
10\,\% & \textbf{0.823} & 0.661 & 0.819 & $+0.159$ \\
25\,\% & 0.826 & 0.752 & \textbf{0.836} & $+0.084$ \\
50\,\% & 0.843 & 0.796 & \textbf{0.854} & $+0.058$ \\
100\,\% & 0.861 & 0.840 & \textbf{0.870} & $+0.030$ \\
\bottomrule
\end{tabular}
\end{table}

\begin{table}[t]
\centering
\caption{Per-class AP (\%) and macro-AP ($\pm$ standard deviation across folds) of
the strategies of Table~\ref{tab:seven} (frozen encoder, head only, 100
epochs; full fine-tuning, 15 epochs) and of the expert features (probe of
Table~\ref{tab:transfer}); 811 spectra, 7 classes, $5{\times}3$
cross-validation. Best value per row in bold.}
\label{tab:7classes}
\vspace{2pt}
\small
\setlength{\tabcolsep}{2pt}\begin{tabular}{lccc}
\toprule
Class & Expert feat. & SSL frozen & \textbf{SSL fine-tuned} \\
\midrule
H$\alpha$ & 94.6{$\pm$1.7} & 94.9{$\pm$2.0} & \textbf{95.7}{$\pm$2.3} \\
molecular & 92.0{$\pm$2.5} & 92.8{$\pm$2.7} & \textbf{93.0}{$\pm$1.8} \\
low-$e^-$ & 84.8{$\pm$3.9} & \textbf{84.9}{$\pm$4.5} & 84.2{$\pm$5.0} \\
cloudy & 89.1{$\pm$3.4} & 89.2{$\pm$2.2} & \textbf{90.5}{$\pm$1.6} \\
dawn/dusk & 64.2{$\pm$25.4} & \textbf{68.4}{$\pm$25.4} & 67.1{$\pm$23.5} \\
{[N\,I]}\,520 & 85.1{$\pm$3.2} & 85.6{$\pm$3.4} & \textbf{86.0}{$\pm$2.6} \\
N$_2^+$\,471 & 92.5{$\pm$2.5} & \textbf{93.0}{$\pm$2.1} & 92.8{$\pm$2.5} \\
\midrule
macro-AP & 86.1{$\pm$3.2} & 87.0{$\pm$3.4} & \textbf{87.0}{$\pm$3.2} \\
\bottomrule
\end{tabular}
\end{table}

Table~\ref{tab:seven} (top) compares fine-tuning strategies on the seven
classes. The pretrained model reaches 0.870 $\pm$ 0.032 mAP (0.804 F1,
0.909 AUROC), against 0.840 for the same model trained from scratch
($+0.030$; 14 of 15 folds, $p=0.001$) and 0.861 for the expert features
($+0.010$; 10 of 15 folds, $p=0.11$, within the dispersion across folds).
Keeping the encoder frozen and training only the linear head gives the
same 0.870 $\pm$ 0.034: the information is already in the pretrained
encoder, and fine-tuning adds nothing essential ($+0.007$ over the frozen
probe, $p=0.15$). With an untrained encoder the same head reaches only 0.732, and the
architecture trained from scratch 0.840: without pretraining the deep model
stays significantly below the expert features (2 of 15 folds, $p=0.003$);
with it, it reaches them. With 100
epochs instead of 15, mAP drops to 0.839, the from-scratch level:
prolonged fine-tuning on 811 labels overwrites the pretrained representation. Per class (Table~\ref{tab:7classes}), AP ranges
from 95.7 (\texttt{ha}) to 67.1 $\pm$ 23.5 for \texttt{dawn\_dusk}, whose 27 positives leave about five per test fold: scarcity of the
class, not a limitation of the model. Against the expert features, no per-class
difference exceeds the dispersion across folds.

Pretraining matters most when labels are scarce, which is the normal regime in this domain.
Table~\ref{tab:seven} (bottom) repeats fine-tuning
with a fraction of the training labels, the expert features being probed
on the same subsets: the gain of the pretrained model over the same model
from scratch grows from $+0.030$ with all labels to $+0.159$ with 10\,\%
(0.819 vs.\ 0.661; 15 of 15 folds), while the pretrained model and the
expert features stay within dispersion at every fraction (0.819 vs.\ 0.823
at 10\,\%, $p=0.72$). Pretraining does not surpass expert knowledge, and combining the two brings no further gain (Table~\ref{tab:transfer}): it brings a model without expert features to the level of one with them.

\subsection{Masking ablation}
\label{ssec:mask}
All masking variants, uniform or informed, at rates 0.35 to 0.75, score
within [0.856, 0.864] on the frozen probe, with a standard error of about
0.010: the representation is insensitive to the masking regime. The only
difference at the edge of significance is the penalty of uniform masking
at 0.75 ($-0.005$; 4 of 15 folds, $p=0.05$), which the informed prior
removes (0.864, the level of uniform masking at 0.35). The benefit of the
prior is therefore time: the encoder only processes visible tokens, so
masking 75\,\% of them costs 70\,min of pretraining instead of 91\,min
for uniform 0.35 when both are timed on the same GPU, 23\,\% less for the
same quality. The two
timed runs also replicate the originals (0.868 and 0.867; $p=0.28$ and
$0.09$): run-to-run variability, about 0.005, is below every difference
reported here.

\subsection{Physical content of the representation}
The encoder received no label, no line catalogue and no expert feature,
yet its frozen output encodes the auroral physics: a ridge regression
($\alpha{=}1$) from the 512-value embedding recovers the five line ratios of the expert
features with a mean R$^2$ of 0.91, against 0.77 for the untrained control
(Fig.~\ref{fig:interp}b). An untrained encoder already preserves much of
this information, so the gap to the control, not the absolute R$^2$,
measures what pretraining learned. The gap is largest on the 636.4/630 control ratio, physically
nearly constant and therefore the hardest to read (0.72 vs.\ 0.41).

To see which wavelengths the fine-tuned classifier actually uses, we
attribute each decision to the input with Integrated
Gradients~\cite{sundararajan2017ig}, integrated from a line-free reference
spectrum to the observed one (64 steps; completeness $\pm0.0\,\%$). Figure~\ref{fig:interp}c gives, per class, the share of positive attribution per spectral window. For \texttt{elow}, 56\,\% of the attribution falls on
the red [O I] doublet, 630.0 and 636.4\,nm (51\,\% on 630.0 alone): the
classifier decides on the high-altitude emission that defines this class.
For \texttt{n2\_471}, the classifier uses both bands
of the N$_2^+$ first-negative system, 427.8\,nm (18.2\,\%) and 470.9\,nm
(16.7\,\%), whereas the annotator inspects only 470.9 and the expert
features measure only 427.8. Their intensity ratio is nearly constant over
the 267 spectra of this class (470.9/427.8 median 0.18, IQR 0.17--0.20), so 427.8 is present whenever the annotator marks 470.9 and
the labels agree; but the classifier has learned the
emitting system rather than the single band a human checks. For \texttt{molecular}, the classifier relies on the N$_2^+$ first-negative
bands at 427.8 and 470.9\,nm (31\,\%) rather than on the N$_2$ first-positive
band the annotator inspects (4\,\%): a correlated signature of the same
molecular-nitrogen emission, not the annotation criterion.

\section{Conclusion}
We pretrained a ViT-1D with a masked autoencoder on 223\,000
unlabelled auroral spectra. The representation captures the underlying
physics without any label and, under one linear probe, classifies as well as 13 expert-designed
features;
fine-tuned, the model outperforms the previous supervised classifier and,
at every label budget, the same model trained from scratch, and
attribution shows that it relies on the whole N$_2^+$ emission system
rather than on the single band an expert checks. None of the three foundation models tested replaces in-domain pretraining, and their transfer
tracks spectral-window overlap: a testable prediction rather than a proven
law, left to a study across instruments.

The representation is held in the encoder and is most useful when labels are scarce ($+0.159$ at
10\,\%); masking rate and an informed static prior change only the
pretraining time.

\noindent\textbf{Acknowledgements.} A large language model (Claude, Anthropic)
helped improve the wording; the authors checked every sentence, and all
methods, experiments, results and figures are their own.

\section{Compliance with Ethical Standards}
This study uses only instrument data (auroral spectra) and involves no human
or animal subjects; no ethical approval was required. No funding was received
for conducting this study. The authors declare no conflict of interest.

\bibliographystyle{IEEEbib}
\bibliography{refs}
\end{document}